# Northeastern Uni at Multilingual Counterspeech Generation: Enhancing Counter Speech Generation with LLM Alignment through Direct Preference Optimization

**Sahil Wadhwa[1,†], Chengtian Xu[2], Haoming Chen[2], Aakash Mahalingam[2], Akankshya Kar[2,‡], Divya Chaudhary[2]**

[1]Capital One, [2]Northeastern University, [3]Apple Inc.

sahil24wadhw@gmail.com, xu.chengt@northeastern.edu, chen.haoming@northeastern.edu, d.chaudhary@northeastern.edu

## Abstract

The automatic generation of counter-speech (CS) is a critical strategy for addressing hate speech by providing constructive and informed responses. However, existing methods often fail to generate high-quality, impactful, and scalable CS, particularly across diverse linguistic contexts. In this paper, we propose a novel methodology to enhance CS generation by aligning Large Language Models (LLMs) using Supervised Fine-Tuning (SFT) and Direct Preference Optimization (DPO). Our approach leverages DPO to align LLM outputs with human preferences, ensuring contextually appropriate and linguistically adaptable responses. Additionally, we incorporate knowledge grounding to enhance the factual accuracy and relevance of generated CS. Experimental results demonstrate that DPO-aligned models significantly outperform SFT baselines on CS benchmarks while scaling effectively to multiple languages. These findings highlight the potential of preference-based alignment techniques to advance CS generation across varied linguistic settings. The model supervision and alignment is done in English and the same model is used for reporting metrics across other languages like Basque, Italian, and Spanish.

Content Warning: This paper includes sensitive content that could be distressing to certain readers.

| Category | Content |
|---|---|
| **HS** | Islam and ISIS should go to hell. All Muslims should be immediately sent to their country because they are all intolerant criminals. If we do so, Britain will be a safer place. |
| **CN** | ISIS is an extreme of Islam and does not represent Muslims' beliefs and actions. Indeed, most of Islam condemned the actions of ISIS. |
| **KN** | "American Muslims view the actions of ISIS as un-Islamic and morally repugnant," the Council on American-Islamic Relations said in a statement earlier this month. It also fits into a clear historic pattern of Western perceptions of Islam and Muslims. We Muslims condemn ISIS's horrible actions. These terrorists, ISIS, are not true Muslims. |
| **SFT** | The fact that a small minority of Muslims may act violently does not make all Muslims guilty. |
| **DPO** | The 'many' extremists are not Muslims; in fact, the majority of ISIS are not Muslims. |

Table 1: Comparison of responses highlighting DPO's advantage over SFT for CN generation.

## 1 Introduction

Traditional methods for tackling hate speech, particularly on social media platforms, have predominantly relied on removing offensive posts or banning users to deter future violations. While these measures can be effective in the short term, they often lead to unintended consequences.

For example, banned users may create new accounts to continue spreading harmful content, perpetuating the problem. Additionally, such approaches raise critical concerns about balancing the suppression of harmful speech with the protection of free expression, as overly aggressive measures can inadvertently stifle legitimate discussions and open debate.

In contrast, Counter Narrative (CN) strategies offer a more constructive and nuanced approach (Schieb and Preuss, 2016), (Kiritchenko et al.,

---

2021). Rather than silencing harmful content, CNs respond with informed, fact-based narratives designed to challenge prejudices and misinformation while promoting understanding. These responses aim to de-escalate hostility and encourage healthier dialogue, directly addressing the biases or misconceptions fueling hate speech. Research has shown that CNs can be effective in reducing the impact of hate speech, fostering more productive online interactions, and mitigating polarization, making them an increasingly compelling option in the fight against hate speech.

However, implementing Counter Narratives (CNs) at scale poses significant challenges. Off-the-shelf Large Language Models (LLMs) often produce generic responses that fail to address the nuanced cultural and contextual factors necessary for effectively tackling diverse hate speech scenarios. This is where Direct Preference Optimization (DPO) can play a crucial role. By refining and aligning LLM outputs, DPO enables the generation of more context-aware, culturally sensitive, and impactful counter-narratives, making it a promising approach for addressing the limitations of traditional LLMs in combating hate speech.

In Table 1, it can be observed that DPO model's response is better than the SFT counterpart in generating counter-speech (CS) that is more contextually relevant and aligned with the grounded knowledge (KN). The DPO model output explicitly refutes the hate speech (HS) by disassociating Muslims from ISIS, stating that "The 'many' extremists are not Muslims, in fact the majority of ISIS are not Muslims," which directly aligns with the KN that condemns ISIS and emphasizes that they do not represent true Islamic teachings. In contrast, the SFT model provides a more generic response, "The fact that a small minority of Muslims may act violently does not make all Muslims guilty," which, while valid, does not leverage the KN effectively to address the specific accusations in the HS.

**Note** - In this paper, we use the terms Counter Narratives (CN) and Counter Speech (CS) interchangeably.

## 2 Related Work

Hate Speech in the past has been tackled in multiple ways. Some works have focused on hope speech, which tackles HS with a constructive view (Palakodety et al., 2020), (Chakravarthi, 2020), and (Ureña López et al., 2023). However, unlike a CN, hope speech does not directly respond to hate speech or counter a message in opposition. (Bonaldi et al., 2024) compares different strategies for tackling hate speech like counter-trolling, anti-stereotyping (Mun et al., 2023), detoxification (Laugier et al., 2021) and misinformation countering (Stammbach and Ash, 2020-10). Each of these methods has its own merits and demerits, but for the scope of this task, we focus on CN generation. Different CN generation strategies have been explored. Constraint-based CN generation leverages various linguistic (Horawalavithana et al., 2022), (Wang et al., 2021), and outcome constraints (Hong et al., 2024) to guide the generation of text. With the advent of Large Language Models (LLMs), there has been a paradigm shift towards leveraging these models for constraint-based counter-narrative (CN) generation, as they don't require prior knowledge of fixed templates or rigid rule sets. LLMs can dynamically adapt to context and generate a wide variety of responses, offering greater flexibility than traditional constraint-based methods. Studies have demonstrated that LLMs, when fine-tuned on hate speech and counter-speech datasets, can produce more contextually relevant and diverse responses. For instance, research by (Saha et al., 2024) evaluated the zero-shot capabilities of models like GPT-2, DialoGPT, ChatGPT, and FlanT5 in generating counter-speech, highlighting the potential and limitations of LLMs in this domain.

Previous studies, such as (Zellers et al., 2019) and (Solaiman et al., 2019), have highlighted that Large Language Models (LLMs) often hallucinate when they lack sufficient context. For instance, early methods focused on predefined responses or templates, limiting their flexibility and scalability. Supervised learning models, while more adaptable, require extensive labeled datasets, which are challenging to obtain for the diverse manifestations of hate speech. These limitations have prompted the exploration of more sophisticated techniques, such as leveraging large language models and reinforcement learning, to enhance the effectiveness and adaptability of CS generation (Hengle et al., 2024). Research has demonstrated that incorporating an external grounded knowledge base significantly enhances the generation capabilities of both conversational agents and LLMs. For conversational agents, grounding responses in external knowledge leads to more accurate, contextually relevant, and fact-based outputs, as shown in studies like (He et al., 2017) and (Dinan et al., 2019). Similarly,

LLMs benefit from this approach by reducing hallucinations and producing coherent and informed responses, as emphasized by (Chung et al., 2021). However, LLMs trained on vast datasets often acquire undesirable biases and attributes, which can be mitigated through human alignment techniques such as Reinforcement Learning from Human Feedback (RLHF) (Ouyang et al., 2022) and Direct Preference Optimization (DPO) (Rafailov et al., 2023). In this paper, we are the first to investigate the effectiveness of model alignment approaches, particularly Direct Preference Optimization (DPO), for generating Counter Speech (CS) to address Hate Speech (HS). By leveraging alignment techniques and grounded knowledge, we aim to improve the quality and relevance of CS generation, enabling LLMs to produce more impactful and scalable responses across diverse linguistic contexts.

This investigation is important because traditional approaches to countering hate speech often fall short in adapting to the nuances of varied cultural and linguistic contexts. Hate speech manifests differently across regions, requiring CS responses that are both context-aware and culturally sensitive. Moreover, the multilingual capabilities of DPO make it especially valuable in addressing hate speech globally, as it allows for the generation of effective counter-narratives across multiple languages. This multilingual usefulness ensures that diverse communities can be supported with relevant and culturally appropriate counter-speech, enhancing the inclusivity and accessibility of digital platforms.

## 3 Dataset

We used the multilingual dataset[*] provided for the shared task as shown in Table 2. The Hate Speech (HS) examples are sourced from the MTCONAN dataset[†], while the Counter Narratives (CN) are newly generated. Additionally, each HS-CN pair is accompanied by five background knowledge sentences, some of which are specifically curated to provide relevant context for generating the Counter Narratives.

We did not use any external dataset for this shared task besides the one in the shared task.

---

[*]https://huggingface.co/datasets/LanD-FBK/ML_MTCONAN_KN

[†]https://github.com/marcoguerini/CONAN/tree/master/Multitarget-CONAN

## 4 Architecture

### 4.1 Pre-trained Models

In this shared task, we leveraged DPO on the Llama-3 (Dubey et al., 2024) model to generate Counter Speech (CS) and demonstrated its superiority over the SFT-only model. We selected Llama-3 as our base model due to its proven effectiveness across multiple NLP benchmarks[‡]. While we also experimented with smaller fine-tuned models like GPT-2 (Radford et al., 2019) and Llama-2 (Touvron et al., 2023), their performance was found to be inferior compared to Llama-3.

### 4.2 Generating Rejected Answers

We optimized LLMs using DPO, leveraging their SFT counterparts as a reference to guide preference-based alignment. Rejected CS responses, as illustrated in Figure 1, were generated using GPT-4o (OpenAI, 2023) to ensure diversity and contextual relevance. The quality of these rejected responses is directly proportional to the quality of the HS. Thus, a low-quality HS would result in a low quality rejected response. These rejected responses were utilized as negative samples in conjunction with preferred responses to fine-tune the LLMs through DPO alignment, improving the quality of generated CS and enabling scalability across diverse linguistic contexts.

In the context of **Counter-Narrative (CN) generation**, rejected answers serve two critical purposes:

- **Defining Negative Samples for Learning:** Rejected answers act as negative examples that help the model understand what constitutes a less-effective or less-preferred counter-narrative. These rejected responses might lack relevance, contextual accuracy, or the necessary persuasive tone to effectively counter hate speech, making them valuable for contrastive learning.

- **Reinforcing Desirable Counter-Narrative Behavior:** By contrasting rejected answers with ground-truth (preferred) counter-narratives, DPO trains the model to prioritize generating responses that are more contextually appropriate, impactful, and aligned with human preferences. This process helps the model learn to avoid unpersuasive, factually

---

[‡]https://github.com/meta-Llama/Llama3/blob/main/eval_details.md

| Split | Number of Examples | Percentage (%) |
| --- | --- | --- |
| Train | 396 | 66.4 |
| Validation | 100 | 16.8 |
| Test | 100 | 16.8 |

Table 2: Data distribution across splits for each language.

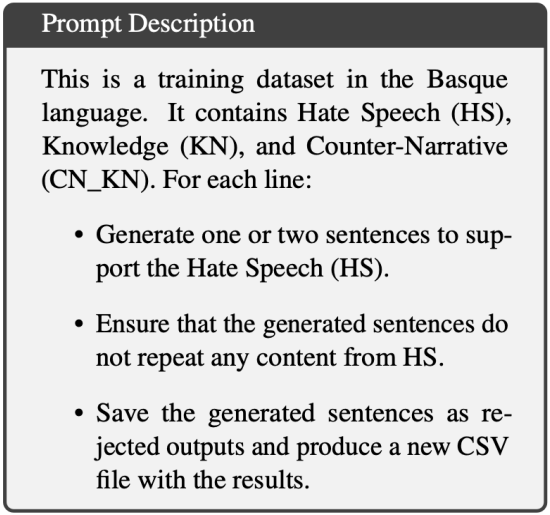

Prompt Description

This is a training dataset in the Basque language. It contains Hate Speech (HS), Knowledge (KN), and Counter-Narrative (CN_KN). For each line:

- Generate one or two sentences to support the Hate Speech (HS).

- Ensure that the generated sentences do not repeat any content from HS.

- Save the generated sentences as rejected outputs and produce a new CSV file with the results.

Figure 1: Prompt description used for generating rejected answers for DPO.

incorrect, or generic counter-narratives while focusing on generating precise, ethical, and contextually rich responses to hate speech.

## 5 Experimental Results

All training processes in this paper were executed on a single 32 GB V-100 GPU. Initially, we applied supervised fine-tuning using the Llama3 basic and instruct models, utilizing default parameters and LoRA (Hu et al., 2022) fine-tuning techniques. The default parameters included a batch size of 4, combining gradients over 4 steps, and weight decay of 0.01. For LoRA, we set the rank (r) to 16, the scaling factor (alpha) to 16, and applied a dropout of 0 to the low-rank layers, targeting the attention layers. The training dataset provided in the shared task was relatively small, consisting of only 1,500 lines, necessitating a higher number of epochs to sufficiently train the SFT model. To prevent excessively long outputs, we set the maximum sequence length to 640. We employed the Adam optimizer with a learning rate of 2e-4, conducting training for 500 epochs for each model. The entire training process spanned approximately 70 hours. After evaluating the models on the validation dataset, we selected the checkpoints at 150 epochs for the Llama3 basic model and 200 epochs for the Llama3 instruct model, referred to as **run1** and **run2** respectively.

Next, we extended the training on our DPO dataset based on the SFT checkpoint. For this phase, we adjusted the learning rate to 5e-4 and continued for an additional 80 epochs for each model. Upon further validation testing, we observed some improvements in the basic model, while the instruct model showed signs of degradation. Finally, we opted for the 80 epochs checkpoint of the Llama3 basic model as our **run3**.

The overall comparison across runs can be seen in Table 3. We provide a detailed evaluation of the models across various metrics to measure their performance in Counter Speech (CS) generation tasks. The metrics used include **AVG BLEU-2** (Papineni et al., 2002), **BERTScore** (Zhang et al., 2020), **JudgeLM** (Zubiaga et al., 2024), and **AVG ROUGE-L** (Lin, 2004). These metrics assess the quality of the generated outputs by measuring their similarity to ground-truth counterspeech, with higher values indicating better alignment with reference texts. Our results show that **run3**, the DPO-aligned Llama3 base model, outperforms the other runs across all metrics, followed by **run2** (SFT Llama3 instruct model) and **run1** (SFT Llama3 base model).

The findings highlight several key lessons learned. First, the superior performance of **run3** reinforces the efficacy of **Direct Preference Optimization (DPO)** for improving text generation tasks, including Counter Narrative (CN) generation. By fine-tuning models with human-aligned preferences, DPO enables outputs that are not only factually accurate but also more assertive and contextually relevant. Second, the comparison underscores the limitations of standard supervised fine-tuning (SFT), which, while effective in generating coherent text, often fails to directly challenge and dismantle hate speech in a targeted manner. Lastly, the integration of metrics such as BERTScore and JudgeLM provides deeper insights into how models align with the subtleties of counter-narrative generation, emphasizing the importance of using diverse evaluation methods. These results demonstrate the value of alignment techniques in producing impactful counter-speech and set a benchmark for future research in this domain.

| Language | Model Name | JudgeLM | RougeL (%) | BLEU (%) | BERTScore (%) | Gen Len | Novelty (%) |
|---|---|---|---|---|---|---|---|
| English | Northeastern Uni run1 | 965.5 | 48.3 | 40.1 | 81.0 | 30.4 | 76.8 |
| | Northeastern Uni run2 | 990.0 | 51.6 | 42.1 | 82.3 | 30.9 | 76.6 |
| | Northeastern Uni run3 | **1191.0** | **51.8** | **40.3** | **82.6** | **43.0** | **78.1** |
| Basque | Northeastern Uni run1 | 1107.5 | 25.6 | 13.3 | 74.6 | 24.8 | 84.3 |
| | Northeastern Uni run2 | **1158.0** | 27.6 | 13.5 | 75.7 | 24.5 | 83.4 |
| | Northeastern Uni run3 | 1145.0 | **30.9** | **17.6** | **76.2** | **29.6** | **85.2** |
| Italian | Northeastern Uni run1 | 830.0 | 42.6 | 30.8 | 79.7 | 32.0 | 77.8 |
| | Northeastern Uni run2 | 905.5 | 45.4 | 33.7 | 80.8 | 33.5 | 76.9 |
| | Northeastern Uni run3 | **1004.0** | **47.5** | **36.2** | **81.3** | **40.7** | 77.8 |
| Spanish | Northeastern Uni run1 | **894.5** | 45.6 | **34.5** | 80.6 | 35.1 | 74.0 |
| | Northeastern Uni run2 | 845.0 | **46.7** | 33.6 | **81.2** | 33.4 | 73.9 |
| | Northeastern Uni run3 | 873.0 | 45.3 | 33.4 | 80.5 | **43.8** | **76.6** |

Table 3: Performance metrics of Northeastern Uni models across languages. The highest scores for each metric are marked in bold

## 6 Error Analysis

Below is an example output generated by the three models:

- **Hate Speech:** Jews are less than animals.

- **Ground Knowledge:** Religion holds up the ideal of the equal dignity of human beings. Why are Americans so ignorant and uneducated about other cultures and religions? Every religion holds up the ideal of the equal dignity of human beings. Teach the human race the futility of attempting to define an unfathomable supreme entity. Why are Americans so ignorant and uneducated about other cultures and religions? Big question.

- **run1 output: (sft Llama3 base model)** Every religion holds up the ideal of the equal dignity of human beings. No matter what, human race must be taught the futility of attempting to define an unfathomable supreme entity.

- **run2 output: (sft Llama3 instruct model)** Every religion holds up the ideal of the equal dignity of human beings. We should all aim to teach the human race the futility of attempting to define an unfathomable supreme entity.

- **run3 output: (dpo Llama3 base model)** Every religion holds up the ideal of the equal dignity of human beings. Are you saying that you are not aware of it and you should teach the human race the futility of attempting to define an unfathomable supreme entity?

Comparing the three outputs, run1 and run2 emphasize education and collective responsibility, suggesting a more generalized and neutral approach to countering hate speech. Both outputs highlight the importance of universal dignity and teaching the futility of defining an unfathomable supreme entity. However, their responses lack specificity in addressing the explicit stereotype presented in the hate speech. The tone remains passive and non-confrontational, making them less impactful in directly opposing the harmful statement. While these outputs might be effective for audiences that are neutral or uninformed, they fail to actively challenge the hateful perspective, potentially limiting their ability to provoke meaningful reflection or change.

In contrast, the output from run3, generated by the DPO-aligned model, adopts a more assertive and interrogative stance. By directly questioning the ignorance implied in the hate speech, it actively confronts the harmful viewpoint and forces the reader to reconsider their stance. This approach, grounded in factual knowledge, provides a stronger rebuttal and creates an opportunity for cognitive dissonance. It balances politeness with firmness, making it more effective in counter-narrative scenarios where directly opposing hate speech is critical. This comparison underscores the importance of fine-tuning with alignment techniques, such as DPO, to produce counter-narratives that are not only coherent but also impactful and assertive in dismantling hate speech.

### 6.1 Future Improvements

The generation of rejected outputs in this work relied on a simple prompt instructing the model to create sentences supporting the Hate Speech (HS) without repeating its content. While this approach served its purpose, the simplicity of the prompt limited the diversity and contextual richness of the rejected outputs. Future improvements could focus on designing more advanced prompts or leveraging techniques such as reinforcement learning to produce more varied and representative outputs. This would enhance the dataset's robustness and support a more comprehensive evaluation of Counter-Narrative (CN) generation models.

We were unable to utilize some of the latest and larger models, such as GPT-4 and certain variants of LLaMA, primarily due to their substantial computational and memory requirements, which exceeded the available resources. Additionally, several state-of-the-art models are not open-sourced, limiting their accessibility for integration into this work. Addressing these constraints in future research could enable the exploration of these powerful models for more advanced and scalable Counter-Narrative (CN) generation.

Additionally, the criteria for rejecting outputs, while necessary for ensuring quality, were somewhat rigid and manual in nature. This limited the exploration of alternative, potentially valid CNs that did not conform strictly to predefined guidelines. Future work could focus on refining rejection criteria to allow for greater nuance and diversity in acceptable CNs, perhaps through semi-automated methods involving human-in-the-loop systems or adaptive scoring mechanisms. By integrating such methodologies, the evaluation process could better capture a broader range of valid CNs, leading to richer datasets and more nuanced performance assessments of CN generation models.

## 7 Conclusion

The integration of Direct Preference Optimization (DPO) has significantly enhanced the generative capabilities of Llama3 models, refining alignment and improving output quality through iterative training on the DPO dataset. This process enables the model to produce contextually relevant and human-preferred outputs, showcasing its potential for impactful applications. While some model configurations demonstrated varying degrees of improvement, the results overall highlight the effectiveness of DPO in driving better performance. These findings underline the value of alignment techniques like DPO in advancing generative models and set the stage for further innovations to unlock even greater potential across diverse applications.

## A Appendix

### A.1 Training Prompt

In this paper, two versions of the LLaMA-3 model were employed. Each model utilized a distinct prompt format during training.

**Prompt used while training LLaMA-3-base:**

Below contains a HS (hate speech) and a KN (knowledge). Based on the knowledge, write a CN (counter narrative) to contradict the hate speech within 50 words.
### HS:
Do you have any proof that the Holocaust ever happened?
### KN:
The ninth is Holocaust Against Jews is a Total Lie - Proof. Yet for the Holocaust survivors and witnesses, the trial is clearly about more than that. The third result is the article The Holocaust Hoax; IT NEVER HAPPENED. At the Eichmann trial close to 100 witnesses testified about their suffering. It has changed both the questions being asked - did the Holocaust actually happen?
### CN:
At the Eichmann trial close to 100 witnesses testified about their suffering. There are living survivos of the Holocaust. What other evidences do you need?<|end_of_text|>

**Prompt used while training LLaMA-3-instruct:**

<|begin_of_text|><|start_header_id|> system <|end_header_id|>
You are an AI assistant. Below contains a HS (hate speech) and a KN (knowledge). Based on the knowledge, write a CN (counter narrative) to contradict the hate speech within 50 words.<|eot_id|><|start_header_id|> user <|end_header_id|>
[HS] Do you have any proof that the Holocaust ever happened? [KN] The ninth is Holocaust Against Jews is a Total Lie - Proof. Yet for the Holocaust survivors and witnesses, the trial is clearly about more than that. The third result is the article The Holocaust Hoax; IT NEVER HAPPENED. At the Eichmann trial close to 100 witnesses testified about their suffering. It has changed both the questions being asked - did the Holocaust actually happen?<|eot_id|>
<|start_header_id|>assistant<|end_header_id|>
[CN] At the Eichmann trial close to 100 witnesses testified about their suffering. There are living survivos of the Holocaust. What other evidences do you need?<|eot_id|>

### A.2 Sample Output

In this section, we present examples from the dataset that illustrate the differences in Counter Narrative (CN) quality generated by the SFT (Supervised Fine-Tuned) model and the DPO (Direct Preference Optimization) model. These examples provide a comparative analysis, demonstrating how the DPO model outperforms the SFT model in terms of factual accuracy, alignment with the ground truth CN, and overall quality.

For example in Table 4, the differences between SFT and DPO model outputs when addressing hate speech. The hate speech in question suggests racial inferiority, while the ground truth counter-narrative (CN) states that no privileges or opportunities should be based on color or origin. The SFT model generates a verbose response that incorporates irrelevant details from the knowledge source, diluting the clarity and focus of the counter-narrative. In contrast, the DPO model produces a concise and directly relevant response, effectively aligning with the ground truth CN by stating, "No jobs should be given to anyone because of their colour or origin." This demonstrates the DPO model's superior ability to filter extraneous information and generate precise, impactful counter-narratives.

| Category | Content |
|---|---|
| HS | Niggas are not as British as we are. They are just an inferior type of citizens. |
| CN | No privilege should be given to anyone because of their colour or origin. |
| KN | Aleksandrovics Lovegrove said nurses would carry around a swatch of colours with different skin tones. She is a fierce fan of British brands and is always proud to 'wear British' both at home and abroad. 2015Mattdavelewis Tom-Felton thoughts on woman of colour being cast as Hermione? But there is a type of red lip that will suit you best, and I'm not necessarily talking about a specific colour. No jobs should be given to anyone because of their colour or origin. |
| SFT | Nurses would carry around a swatch of colours with different skin tones. But there is a type of red lip that will suit you best, and I'm not necessarily talking about a specific colour. No jobs should be given to anyone because of their colour or origin. |
| DPO | No jobs should be given to anyone because of their colour or origin. |

Table 4: Sample Outputs for LLaMA 3 base model: DPO is Clearer and Closer to Ground Truth

| Category | Content |
|---|---|
| HS | LGBTs have no place in a civilised society. |
| CN | Countless people, particularly kids, face fear and abuse every day because of their sexual orientation. The majority of states also lack anti-discrimination protection based on sexual orientation or gender identity. We should fight to make our country all for all people with different sexual orientations. |
| KN | Has he not noticed that we hardly have a civilised society? foie gras is not a healthy product and has no place in a civilised society. Countless people, particularly kids, face fear and abuse every day because of their sexual orientation. The majority of states also lack anti-discrimination protection based on sexual orientation or gender identity. Again means Make America all for white people, no immigrants, no people of different sexual orientations. |
| SFT | Countless people, particularly kids, face fear and abuse every day because of their sexual orientation. A tolerant society is not only for the majority but also for the minorities. |
| DPO | Countless people, particularly kids, face fear and abuse every day because of their sexual orientation. The majority of states also lack anti-discrimination protection based on sexual orientation or gender identity. Does this mean we are not a civilised society? |

Table 5: Sample Outputs for LLaMA 3 base model: DPO utilizes more background knowledge

| Category | Content |
|----------|---------|
| **HS** | We cannot be associated with people who are unwilling to integrate. |
| **CN** | So what needs to be done to ease integration for migrants and refugees? Ultimately, successful integration will come down to how much civil society will continue to welcome and support the migrants, ensuring that they don't feel isolated in their new homes. |
| **KN** | So what needs to be done to ease integration for migrants and refugees? Ultimately, the planet's future depends on successfully integrating the economies of rich countries and poorer ones. But, you can manage the extent to which culture detracts from rather than contributes to successful integration. The irony is, the more successful the EU becomes in blocking the migrants, the more get trapped in Morocco. But, he says, successful integration will also come down to how much civil society will continue to welcome and support the migrants, ensuring that they don't feel isolated in their new homes. |
| **SFT** | The extent to which culture detracts from rather than contributes to successful integration can be managed. Additionally, how much civil society will continue to welcome and support the migrants will greatly impact the integration process. |
| **DPO** | Successful integration will also come down to how much civil society will continue to welcome and support the migrants, ensuring that they don't feel isolated in their new homes. |

Table 6: Sample Outputs for LLaMA 3 instruct model: DPO focus on background knowledge